# Too sick for surveillance: Can federal HIV service data improve federal HIV surveillance efforts?

**Nick Williams, Ph.D**
Applied Clinical Informatics Branch
Lister Hill National Center for Biomedical Communications,
National Library of Medicine
Bethesda, Maryland, United States of America
Nick.williams@nih.gov

## Abstract

**Introduction:** The value of integrating federal HIV services data with HIV surveillance is currently unknown. Upstream and complete case capture is essential in preventing future HIV transmission.

**Methods:** This study integrated Ryan White, Social Security Disability Insurance, Medicare, Children's Health Insurance Programs and Medicaid demographic aggregates from 2005-2018 for people living with HIV and compared them with Centers for Disease Control and Prevention HIV surveillance by demographic aggregate. Surveillance Unknown, Service Known (SUSK) candidate aggregates were identified from aggregates where services aggregate volumes exceeded surveillance aggregate volumes. A distribution approach and a deep learning model series were used to identify SUSK candidate aggregates where surveillance cases exceeded services cases in aggregate.

**Results:** Medicare had the most candidate SUSK aggregates. Medicaid may have candidate SUSK aggregates where cases approach parity with surveillance. Deep learning was able to detect candidate SUSK aggregates even where surveillance cases exceed service cases.

**Conclusions:** Integration of CMS case level records with HIV surveillance records can increase case discovery and life course model quality; especially for cases who die after seeking HIV services but before they become surveillance cases. The ethical implications for both the availability and reuse of clinical 'HIV Data' without the knowledge and consent of the persons described remains an opportunity for the development of big data ethics in public health research. Future work should develop big data ethics to support researchers and assure their subjects that information which describes them is not misused.

## 1 Introduction

Case capture qualifies every epidemiological measure in infectious disease models as well as the conclusions a model supports(1). Prevalence and death can be modeled from case capture to produce common measures such as attack rates, SIRD (Susceptible, Infected, Recovered or Dead) and mortality fractions(2–5). The number of cases for a given condition at a point in time is high value for any effort to interrupt infections, prevent injury or save public dollars(6–10).

The Centers for Disease Control and Prevention (CDC) attempts to harmonize known sub-

federal HIV case discovery of persons living with HIV via its National HIV Surveillance System (NHSS). Beyond surveillance, The Human Resources Services Administration's (HRSA) Ryan White Programs (A and B), The Centers for Medicare and Medicaid Services (CMS), Social Security Administration's (SSA) Social Security Disability Insurance (SSDI) and Veterans Affairs (VA) provide direct and indirect HIV services to persons living with HIV. The ability of these federal HIV service efforts to inform federal HIV surveillance over time remains unknown.

This study collects publicly available data from several federal HIV service and surveillance efforts and integrates their data with demographic aggregates learned from CMS identifiable claims data. The goal of this study is to find opportunities to improve federal HIV surveillance using existing federal HIV services data. Service demographic aggregate volumes which exceed surveillance volumes are opportunities for sharper case capture, as cases are known to service providers but not surveillance programs. If any federal service program exceeds surveillance, then surveillance is incomplete within a demographic aggregate, suggesting that some populations are harder to surveil than others and federal service programs could improve case capture. Deep learning models may be able to identify aggregates where persons living with HIV are known to services and unknown to surveillance even if the surveillance volume exceeds the service volume within aggregate.

## 2      Methods

### 2.1     Surveillance records

This paper considers six federal HIV service programs and one surveillance program. The surveillance program was exclusively captured from the CDC report series, '*Diagnoses of HIV Infection in the United States and Dependent Areas, YYYY'*(11–20). Ten annual publications (which describe different 'data years,' or YYYY) from the series were considered. The series was retrieved from CDC Stacks (https://stacks.cdc.gov/), a hosting server for historic CDC publications. These publications detail multi-year PDF tables of demographic aggregates of people living with HIV known to CDC in the United States (US). The   tables are populated from collections of state specific surveillance reporting sources which can include hospitals, labs and other facilities. These surveillance reporting sources change over time; CDC achieved all states (50) and territories (6) reporting at least some data to CDC in surveillance year 2012. Surveillance data in this study was learned from the most recent publication year available, as some reports contain multiple observation years. Measures of interest were deaths among and prevalence of persons living with HIV; incidence was not considered.

### 2.2     Service records

CMS records provided multiple service program records from Medicare, Medicaid, Child Health Insurance Program(s) (CHIP) and SSDI.  For the purposes of this article, our CMS records should be considered 'identifiable'. Our study was exempted because data did not constitute human subjects research provided the analysis dataset was deidentified; even though its source data was identifiable records. We followed all relevant protocols for transforming identifiable CMS data into the aggregated, deidentified study data set. Do note that surveillance efforts are not research, any more than service beneficiaries are research subjects. In this paper the term 'cases' is used to refer to both, as both are being studied here.

Towards CMS case capture, nearly all Americans will be covered by CMS at some point in the life course; and CMS is a major HIV payor in the United States(21). Medicare Part-D is implicated in Ryan White 'Doughnut Hole' cases who (re)enter the workforce and receive Part-D coverage while waiting for private insurance coverage to start(22,23). Medicare proctors SSDI; after two years of disability coverage SSDI claimants become Medicare eligible independent of age(24,25). CMS claims from 1999-2018 were considered for Medicare and Medicaid programs with Medicare claims providing SSDI services data and

Medicaid containing CHIP services.

Persons living with HIV and seeking services through Medicare who were under the age of 63 at Medicare enrollment were assigned to SSDI services independent of age at observation year. SSDI cases may age through the '65 years old' at enrollment date and appear in observation years with 'traditional' Medicare enrollees, but they are disambiguated below as SSDI independent of age at observation. The Medicare group is also complemented by the '64 years old at enrollment' aggregate which contains people living with HIV who were unlikely to be SSDI cases and were enrolled technically before their 65th birthdays (often by days). Such cases could be age-discordant marital spousal inheritance cases; they were indexed as Medicare 60-64 age aggregate. Distinct deaths were considered within services as well within an artificial cohort, 'CMS', which only contains people living with HIV who had a date of death known to CMS. CMS cohort cases were indexed independent of servicing program.

Within HRSA Ryan White parts A (services) and B (medications) were considered. Ryan White aggregate volumes were discovered through HRSA PDF reports series 'Ryan White HIV/AIDS Program, AIDS drug Assistance Program (ADAP) Annual Client Level Data Report YYYY'(26) and 'Ryan White HIV/AIDS Program Annual Client-Level Data Report, Ryan White HIV/AIDS Program Services Report (RSR)'(27–31). HRSA reports describe retrospective service years within reporting years for a HRSA service range of 2010-2019. Table 1 describes the data sources, the service programs considered, their time ranges and if they were sourced from public documents (PDF) or encounter level records.

**Table 1. Aggregate collection**

| Agency | Program | Range | Vintage |
|---|---|---|---|
| CDC | Surveillance | 2005-2018 | PDF Aggregates |
| HRSA | Ryan White Part A (RSR or RWA) | 2010-2019 | PDF Aggregates |
| HRSA | Ryan White Part B (ADAP or RWB) | 2014-2019 | PDF Aggregates |
| SSA | SSDI via Medicare | 1999-2019 | Encounter Records |
| CMS | Medicare (Total Universe - SSDI) | 1999-2019 | Encounter Records |
| CMS | Medicaid (Total Universe + CHIP) | 1999-2018 | Encounter Records |
| CMS | CMS (Distinct HIV+ Deaths) | 2001-2019 | Encounter Records |

## 2.3    Surveillance completeness

All HIV surveillance record years should be considered inherently incomplete because HIV testing is sporadic, and it can take several latency period years post infection for a patient to present with AIDS clinically. Time to AIDS (in leu of medication), sporadic testing and HIV latency periods should all provide lag functions to surveillance (and services) capture efforts. Further, not all clinically detected subjects are reported to surveillance reporting sources in all study years, which in turn report to CDC. Some persons living with HIV could seek federal HIV services before HIV surveillance capture and die before HIV surveillance capture. Such subjects would be some, though not all, of the Surveillance Unknown, Service Known (SUSK) cases sought in this study.

## 2.4    Services comparison

Comparison between service programs is most likely inappropriate without case reidentification and disambiguation, which was not attempted in this study for several reasons. CMS services data reidentification is banned under our data use agreement and non-CMS services data in this study is natively aggregated to prevent such uses. People living with HIV most likely appear in multiple service data sets; while Medicare and SSDI cases are disambiguated within the study year by design, they are perhaps eligible for Medicaid and are often dual enrolled. Further any Medicare or Medicaid case could also be a Ryan White case. Roughly 40% of Ryan White cases are explicitly covered by Medicare or Medicaid services in a given year in the PDF reports used in this study. Policy changes, multiple payors, eligibility changes and death should all be considered barriers to direct comparison between de-identified service data across services. People living with HIV could also be 'implicitly covered'; where they receive a service at some point in time but do not count as a reportable beneficiary in every data year. In turn, comparing services to services should be undertaken with caution.

## 2.5 HIV attribution

For people living with HIV known to surveillance, any case declared by CDC surveillance reports was considered HIV+. CDC surveillance cases should have a reactive and confirmatory HIV diagnostic test (two different test types) or an AIDS diagnosis to qualify; though HIV case definitions and testing requirements have changed over the study period(32). For Medicare and Medicaid sourced services cases, the Chronic Conditions Warehouse (CCW) definition for HIV was considered proof of infection. The CCW definition is conservative and contains high certainty beneficiaries only. Medicare and Medicaid cases that met the HIV case definition were considered HIV+ in every surveillance year though most likely some (perhaps few) cases seroconverted while a CMS beneficiary. HIV attribution to Ryan White beneficiaries is also complex, as child dependents of people living with HIV who sought services from Ryan White can be covered and do not necessarily have HIV. In turn Ryan White cases under 19 years of age should be interpreted with care. Further, CMS cases within the 65+ group most likely contain people living with HIV who aged into Medicare with HIV as well as individuals who contracted HIV late in life (after age 65). Medicaid provides CHIP observation years for individuals who may not have HIV infection during their CHIP years but are observed aging and reporting HIV services to Medicaid.

## 2.6 Study data set

This study uses a case aggregate index of several demographic measures and compares their case volumes between surveillance and service within aggregates. Aggregate domains considered include Age_Death, Race_Death, Age, Place_Race, Race and Race_Age. Age was aggregated as five-year age groups. Aggregate inclusion was determined by 'PDF' source availability. If available across PDF sources, aggregates were computed from the CMS identifiable records within program attributions. Gender was not considered, as it was not available in all surveillance years. Perhaps one in four Medicare and Medicaid sourced persons living with HIV was declared female at some point over the study period by their respective data sources.

The case aggregates studied here were hand transcribed in the case of PDF's or calculated for patient level identifiable records; both were stored in an integrated long index data model. In each analysis surveillance aggregates were matched to service aggregates by surveillance or service year and aggregate domain. The final index contained 24,753 aggregate records across all surveillance and service programs. Aggregates were harmonized across services and surveillance for 543 distinct kinds of aggregates such as age "65+" or place-race combinations like "Maryland_Black". The volume of an aggregate is the distinct individuals described at a given point in time. Table 2 shows a sample of the index without calculated values.

Service volumes were joined to surveillance volumes by aggregate domain and year. Computed values include: Domain Share, or the percent of the volume within the aggregate program and year, Surveillance Share or the percent of the service volume expressed out of the surveillance volume within an aggregate domain and year; and Surveillance Remainder, or the services volume subtracted from the surveillance volume within aggregate domain and year. The final study data set describes the surveillance program or service program volumes for cases within aggregates for a given year.

**Table2. Index sample values**

| Origin | Program | Year | Domain A | Domain B | Volume | Aggregate |
|---|---|---|---|---|---|---|
| CMS | Medicaid | 2015 | 55-59 | | 31,548 | Age |
| CDC | Surveillance | 2009 | Maryland | Black | 22,858 | Place_Race |
| SSA | SSDI | 2015 | Hispanic | | 15,329 | Race |
| HRSA | RWA | 2014 | White | | 134,363 | Race |

## 2.7 Analysis

For a hypothetical example, if CDC reports 50,000 people in 'Place_Race' within domain 'Maryland-Black' as living with HIV in year 2050 and Medicare reports 70,000 persons living with HIV, with racial demography declared as Black who received mail in the state of Maryland in 2050 one could conclude that Medicare knows of persons living with HIV that CDC does not, specifically within the 'Maryland-Black' and year '2050' domain. This would be a case of SUSK candidacy because service cases exceed surveillance. Because this is an aggregate comparison the share of the services volume that are SUSK cases is unknown. SUSK cases where services exceed surveillance are considered in Table 3 and Figure 1. Table 3 considers these excesses explicitly at aggregate level while Figure 1 considers counts of aggregates containing services exceeding surveillance volume over time.

Simply because surveillance cases exceed services does not mean surveillance has complete services case capture. This is explored in Figure 2 and 3 as well as Tables 4 and 5. Figure 2 considers the distribution of a domain within an aggregate-year and compares services to surveillance. For example, consider if Blacks made up 20% of persons living with HIV in Maryland known to surveillance with an n of 100,000 (.2) and made up 40% of cases known to a service with an n of 50,000 (.4) in the same study year. This hypothetical difference, despite having more surveillance cases, may indicate SUSK cases in the service volume. It could also mean service eligibility, access or program participation bias. Figure 2 plots .2 vs .4, on a scatterplot by program and consider the difference across aggregate domains.

Figure 3 looks for outliers in deep learning model outputs that attempt to guess service volumes within aggregates and domains from surveillance and vise-versa. The statistic of interest is the difference between 'variable importance' which deep learning reports at program, aggregate and domain level. If surveillance and service data is 'complete' services should predict surveillance and surveillance should predict services, because they would be identical, except for service program bias. Deep learning models knew program, year, domains, service, and surveillance volumes. The difference (diff) between the aggregate variable importance scores across models can identify aggregates where surveillance and services differ statistically despite having more surveillance cases. Table four describes the R squares(r2) from the models, where an r2 of 1 would mean complete discovery of all volumes from their alternates. Table 5 lists high (need to know) rankings for aggregates from the models. The model can be used to find gaps in surveillance or gaps in services, where said gaps are regions of the model where correctly guessing a service volume from surveillance, or surveillance volumes from services requires the alternate volumes to guess correctly. The model cannot be used to identify, assign, or target persons living with HIV, but can find aggregates that are underrepresented in services or surveillance volumes.

### 2.7.1 Analysis 1: Finding SUSK volumes when services exceed surveillance

The primary analysis sought to discover excess volumes within programs. This was achieved with the `Surveillance Remainder` measure. A large remainder (positive number) means services are discovering people living with HIV unknown to surveillance so frequently that the service n exceeds the surveillance n. Because the ground truth of identity is unknown between surveillance and services, an aggregate that has more service than surveillance cases is not necessarily completely under-described by surveillance but is at least partly under described. Services cases under described in surveillance within aggregates are detailed in Table 3. Aggregates that exceeded their surveillance share were plotted on a timeline to consider the duration of the error in Figure 1. Only aggregates with a corresponding surveillance aggregate for the given data year were considered.

### 2.7.2 Analysis 2: Finding SUSK volumes when surveillance exceeds services

The Domain Share was used to plot the difference in the distributions of domains within aggregates. Figure 2 describes the domain share as plots by aggregate and index classes. A deep learning model approach was also implemented using h2o.ai(http://www.h2o.ai/), where two volume predictions were made, (surveillance and services) (33).

The deep learning models used in this study are 'multi-layer feed forward neural network' with regression evaluation models (34). In this class of model, the relationship between services and surveillance aggregates is considered by an algorithm which produces information about non-linear relationships between them and then uses a regression model to evaluate the machine generated non-linear relationships. That final regression makes output statistics describing the relationships between services and surveillance volumes. The model is multi-layered because multiple 'hidden' relationship layers between observed facts are generated by a computer. A computer can generate hundreds of complex relationships between two observed facts (a network of neurons) and use relationships between many observed facts and computed relationships to improve conclusions offered about other observed facts. In between these fact to layer to layer to fact relationships an error term is computed iteratively to evaluate the certainty of the relationships between layers and observed facts. For each record (training example or data set row) the algorithm operates and attempts to minimize its loss function, or inaccuracy in its understanding of the observed fact to observed fact relationships that pass through the network of non-linear computed relationships.

The algorithm attempts to minimize this error or misunderstanding from a default understanding of these relationships. It learns deeply, because it attempts to minimize the error several times using information from across the machine generated associations in the neural network. In this study the loss function is being minimized between the services to surveillance relationships and the variable importance in minimizing that error is presented in results section. Feed forward models are models where the path from dependent variable to nonlinear machine generated facts to independent variables is linear, non-recursive and one directional.

The differences in variable importance reported by the models were plotted to detail the service-surveillance incommensurate regions of the index. Only volumes where surveillance was larger than services were considered in the deep learning models. If a variable was of high importance the model 'needs' to know it to produce a meaningful guess of the volume type (surveillance or services). Low importance index features are not meaningless or irrelevant but are perhaps well described across the dimensionality of the index aggregates between both surveillance and services. The difference between the `variable importance surveillance aggregates` and `variable importance services aggregates` indicates the relative difference in model cardinality in aggregate level terms. Because only larger surveillance volume aggregates were considered, outliers in variable importance differences between aggregates indicate aggregates where services have cases unknown to surveillance or surveillance has cases unknown to services.

Deep learning was chosen because it can scale to large data volumes. If future models seek to use sharper geo-specificity (from states to counties or zip codes) its ability to scale to larger data sets is desirable. Deep learning can also evaluate 'why' it made a decision by returning not just the variable/column with a statistical ranking but the item response within the column and its specific rank. Given the data model's use of labeled aggregates, receiving a specific result can inform future surveillance development or services outreach work. The deep learning model variable importance scores are compared below; note the within column item response rankings.

Models were validated across five cross validations with 25% training and 75% evaluation; no model tuning was attempted. Note the goal of the model is not to provide a 'prediction' but to find 'hard to predict' or high importance aggregates. Specifically, the models sought to detect outlier variable importance attributed to aggregate domains, which are described in Figure 3 and Table 5. Only aggregates with a corresponding surveillance aggregate for the given data year were considered; and only volumes where surveillance exceeds services were considered in the second analysis.

### 2.8 Human Participant Compliance Statement

This study complied with all CMS privacy board requirements, and access to CMS data is governed by a comprehensive data use agreement. This study does not constitute human subjects research. This study was exempted from traditional Internal Review Board (IRB) review under exemption category four subsection two: "Exemption Category Four Applies to secondary research of identifiable private information or identifiable biospecimens, if at least one of the following criteria is met: (1) When the identifiable materials are publicly available or (2) when the data is recorded by the investigator in a de-identified manner (analysis data set), i.e. no identifiers are accessible to the research once the analysis begins.  For example, the researcher conducts a retrospective medical chart review and records the necessary data in a datasheet for future analysis without any personal identifiers nor a code which would allow the investigator to link back to subjects."

## 3 Results

### 3.1 Analysis 1: Finding SUSK volumes when services exceed surveillance

Table three presents service aggregates which exceed surveillance and the case volume contained within said aggregates. Aggregates with services overages, and their share of the total is presented.

**Table 3. National services volume excess by aggregate, non-distinct volume and source**

| Program | Class | Aggregates | Exceeding Surveillance | Excess Share | Excess Candidates | Case Volume | Volume Share |
|---|---|---|---|---|---|---|---|
| Medicare | Age_Death | 12 | 12 | 100.00% | 39,629 | 39,629 | 100.00% |
| Medicare | Age | 28 | 10 | 35.71% | 611,149 | 867,809 | 70.42% |
| Medicare | Race_Age | 70 | 33 | 47.14% | 426,284 | 609,474 | 69.94% |
| CMS | Age_Death | 120 | 12 | 10.00% | 55,814 | 128,154 | 43.55% |
| CMS | Race_Death | 65 | 26 | 40.00% | 56,304 | 131,894 | 42.69% |
| Medicaid | Age_Death | 118 | 6 | 5.08% | 10,903 | 60,207 | 18.11% |
| Medicaid | Age | 166 | 4 | 2.41% | 107,109 | 2,526,880 | 4.24% |
| Medicaid | Race_Age | 673 | 50 | 7.43% | 64,498 | 1,565,739 | 4.12% |
| RWA | Race_Age | 364 | 62 | 17.03% | 31,588 | 2,113,393 | 1.49% |
| RWA | Place_Race | 1,750 | 292 | 16.69% | 35,392 | 2,532,144 | 1.40% |
| RWA | Age | 117 | 6 | 5.13% | 42,600 | 4,715,197 | 0.90% |
| Medicaid | Place_Race | 1,925 | 97 | 5.04% | 10,826 | 1,651,595 | 0.66% |
| RWA | Race | 63 | 13 | 20.63% | 19,233 | 4,619,370 | 0.42% |
| SSDI | Race_Age | 494 | 9 | 1.82% | 7,620 | 2,015,888 | 0.38% |
| SSDI | Place_Race | 1,960 | 39 | 1.99% | 2,531 | 2,110,240 | 0.12% |
| RWB | Place_Race | 350 | 25 | 7.14% | 323 | 274,005 | 0.12% |
| Medicaid | Race | 98 | 2 | 2.04% | 902 | 2,171,290 | 0.04% |
| Medicare | Place_Race | 1,300 | 3 | 0.23% | 147 | 632,738 | 0.02% |
| Medicare | Race | 70 | 0 | 0.00% | 0 | 846,704 | 0.00% |
| RWB | Age | 65 | 0 | 0.00% | 0 | 1,347,585 | 0.00% |
| RWB | Race | 35 | 0 | 0.00% | 0 | 1,331,977 | 0.00% |
| RWB | Race_Age | 91 | 0 | 0.00% | 0 | 283,892 | 0.00% |
| SSDI | Age | 151 | 0 | 0.00% | 0 | 2,826,066 | 0.00% |
| SSDI | Age_Death | 120 | 0 | 0.00% | 0 | 85,820 | 0.00% |
| SSDI | Race | 70 | 0 | 0.00% | 0 | 2,778,344 | 0.00% |

Surveillance and service programs change over time. Understanding retrospective vs prospective study value is important. Figure 1 shows changes in services overtaking surveillance as a count of aggregates over time. Note surveillance death data was not available for 2017 onward.

**Figure 1. Service volumes with surveillance excesses over time by aggregate class with local area weighted smoothing**

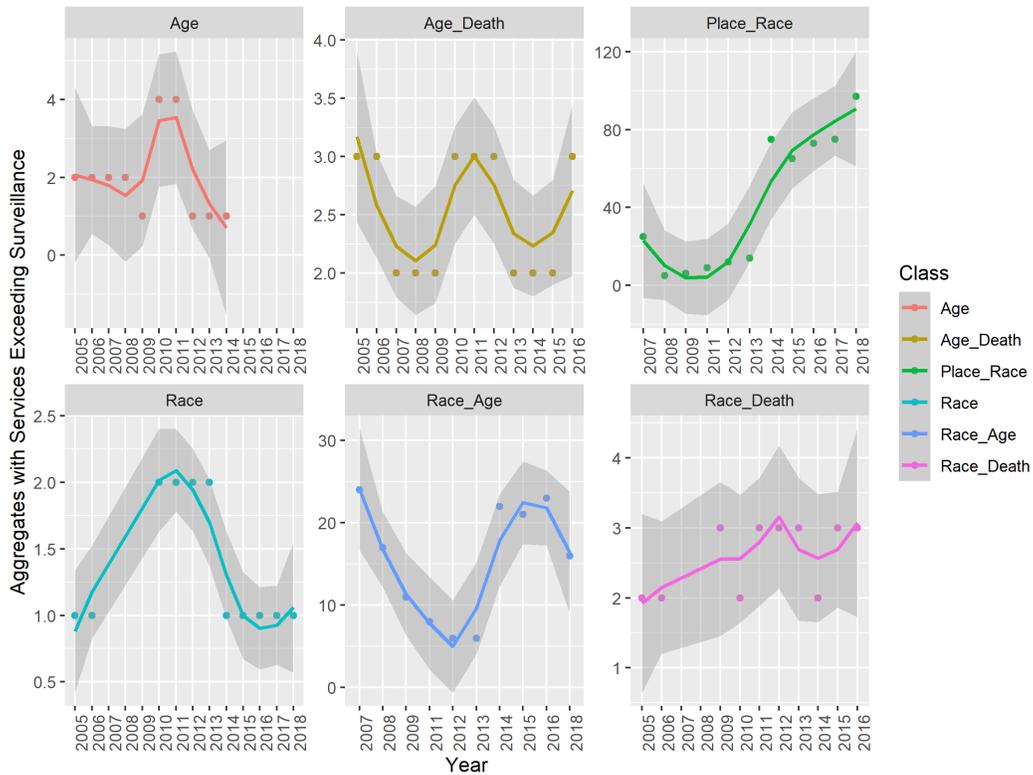

Figure 1: Figure one is a scatterplot matrix where the x axis are the given study year and the y axis are counts of aggregates where services volumes exceeded surveillance. Plots are faceted by aggregate type detailed in the legend.

### 3.2 Analysis 2: Finding SUSK volumes when surveillance exceeds services

Figure 2 plots the share of the service volume within year and aggregate against the share of the same aggregate within services as a scatter plot. Agreement would produce a linear trend, and disagreement would produce a non-linear placement. A jittered placement suggests small disagreement. Local weighted smoothing indicates where aggregates 'should' be plotted assuming their linear similarity.

**Figure 2. Distributions within Aggregates for Domains of Volumes where Surveillance Exceeds Services with Local Area Weighed Smoothing**

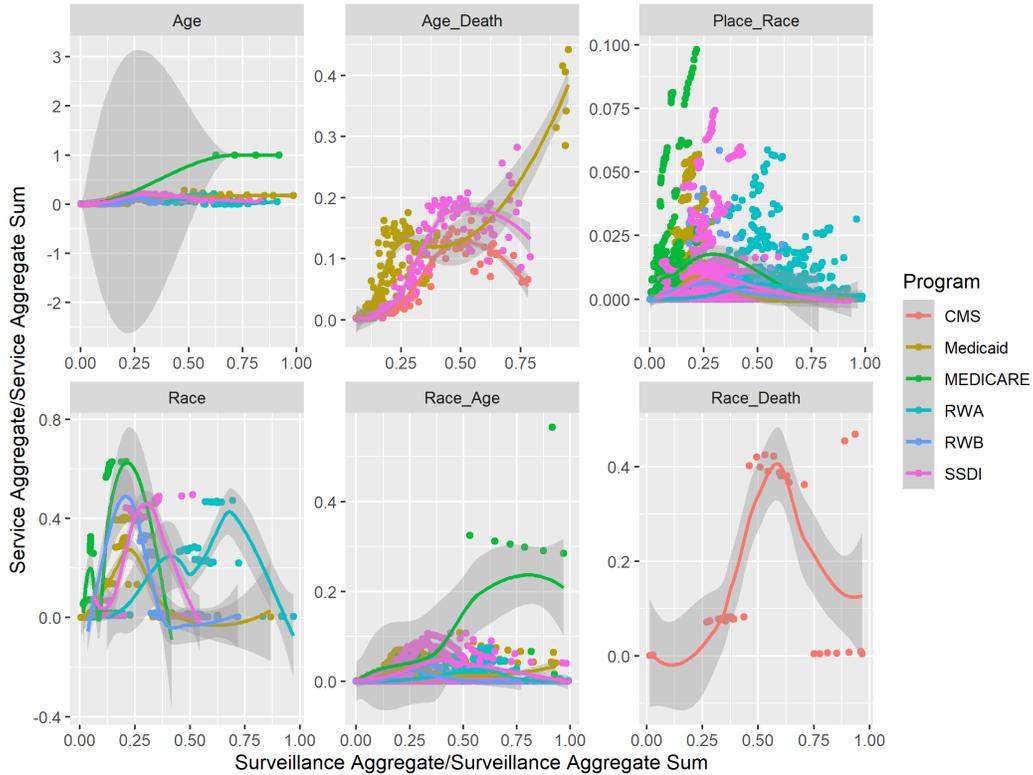

Figure 2: Figure 2 shows a scatter plot matrix where the x axis detail the share of the surveillance aggregate within the sum of the surveillance aggregate, and the y axis shows the share of the service aggregate within the sum of the service aggregate. Each aggregate item response, such as a given race or age group is plotted against the x and y axis. Item responses are colored by program, as described in the legend. Local area weighted smoothing reveals where similarly trending item responses disagree from expectation.

Figure 3 describes the variable importance of two deep learning models. The models attempted to predict surveillance volume within aggregates from services volume and vise-versa. Models only knew services aggregates with superior surveillance volumes. The difference (diff) region in the right line graph should be understood as variable service importance subtracted from variable surveillance importance where zero would mean equal response, a positive number would mean more important for services and a negative number more important to surveillance within an individual aggregate. The X axis is an ordered list of variables/ aggregates present in both models. Origin, or data source was of high importance to understanding the distribution of service volume (blue), and volume class (record type) was of high importance for surveillance (green). Note the red aggregates which depart from the expected variance observed for most aggregate cases (longer red lines); these could be SUSK candidates.

# Figure 3. Deep learning aggregate importance outliers for surveillance exceeding service volumes

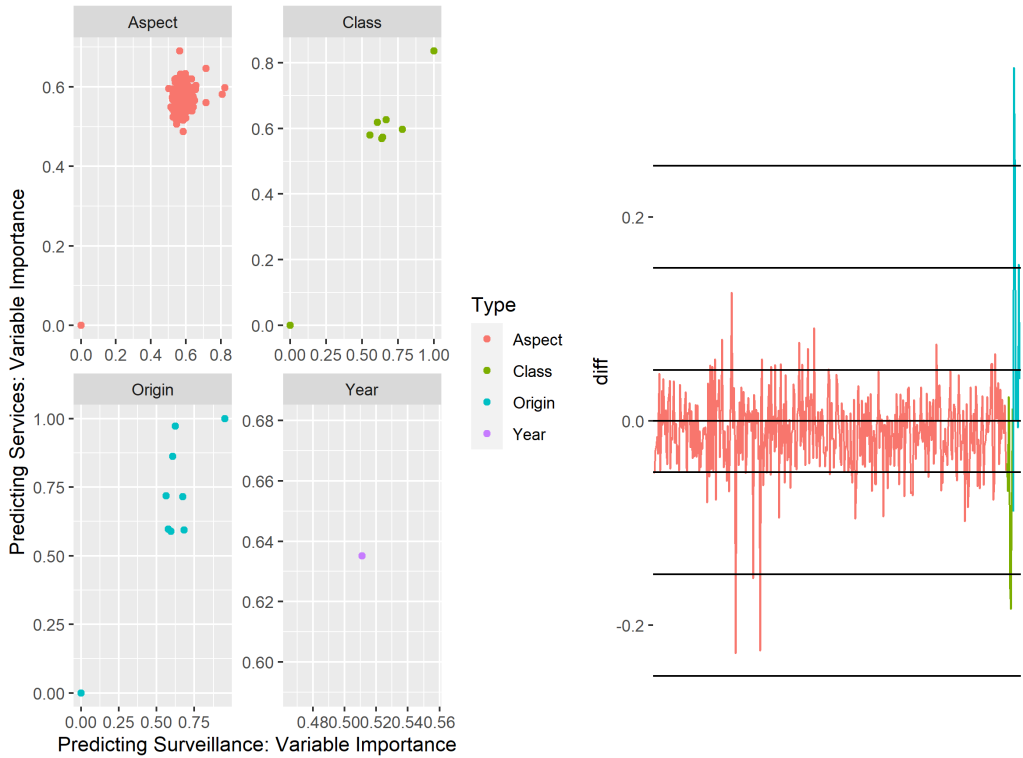

Figure 3: Figure three details the model values from a deep learning model. The right hand side plots the difference in variable importance in predicting services from surveillance, and surveillance from services data. The left figure plots a scatterplot matrix that renders surveillance prediction model variable importance on the x axis and services prediction model variable importance on the y axis. Specific aggregates are plotted and faceted by type of aggregate.

Table four list the model value r2 statistics which describe the accuracy of the model at different stages.

# Table 4. Deep learning model r2

| r2 | Predicting Services | Predicting Surveillance |
|---|---|---|
| Training 25% | 0.878169 | 0.984287 |
| Validation 75% | 0.825476 | 0.969657 |
| Cross Validation | 0.84797 | 0.964012 |

Table five lists the top fifteen most important variables for predicting surveillance or services accompanied by the alternate ranking from the opposite model for ease of comparison. Numerical scores are out of 1, where 1 is the highest and the remaining

rankings are ranked relative to 1.

Table 5. Largest Variable Importance with Alternative Rankings

| Aspect | Predicting Services | Alternate Surveillance | Aspect | Predicting Surveillance | Alternate Services |
|---|---|---|---|---|---|
| Origin.RWA | 1 | 0.9509 | Class.Place_Race | 1 | 0.8358 |
| Origin.Medicaid | 0.9716 | 0.6253 | Origin.RWA | 0.9509 | 1 |
| Origin.MEDICARE | 0.8625 | 0.6061 | aspect.Black | 0.8227 | 0.5979 |
| Class.Place_Race | 0.8358 | 1 | aspect.AIAN | 0.8083 | 0.5809 |
| Origin.SSDI | 0.7183 | 0.5646 | Class.Race | 0.7806 | 0.5967 |
| Program | 0.7147 | 0.6738 | aspect.65 | 0.7154 | 0.6465 |
| aspect.65 Black | 0.6904 | 0.5646 | aspect.Asian | 0.7144 | 0.5602 |
| aspect.65 | 0.6465 | 0.7154 | Origin.CMS | 0.6819 | 0.5933 |
| Year | 0.6352 | 0.5111 | Program | 0.6738 | 0.7147 |
| aspect.FL White | 0.6327 | 0.5966 | Class.Age | 0.6678 | 0.6262 |
| aspect.CA Asian | 0.6317 | 0.571 | aspect.45_49 | 0.6573 | 0.6028 |
| aspect.White | 0.6309 | 0.5743 | aspect.CA Hispanic | 0.6565 | 0.5934 |
| Class.Age | 0.6262 | 0.6678 | aspect.TX Hispanic | 0.6499 | 0.5655 |
| aspect.50-54 White | 0.6234 | 0.5627 | aspect.NC Black | 0.6431 | 0.5496 |
| aspect.25-29 Black | 0.6217 | 0.586 | Class.Race_Age | 0.6425 | 0.5728 |

## 4  Discussion

### 4.1  Findings

The key finding above is that age and race volumes from surveillance are not very different from services despite detecting extra deaths and strong disagreement in Place_ Race. This should mean that services are detecting similar kinds of people (age, race) but are finding different people living with HIV relative to Place_Race residency and death. Further, the error is complete for the Medicare population deaths, suggesting that these SUSK cases die before surveillance can detect them (Table 3). The service death volumes are of special concern, as deceased persons living with HIV who are unknown to surveillance have become infected, potentially transmitted and died without informing surveillance models or surveillance conclusions.

Medicare Age at Death has the most consistent excess volume error at 12/12 followed by Medicare 10/28 (Table 3). Excess service volumes must contain distinct people living with HIV known to Medicare and not to CDC. Medicare age also indicates, even after removing the SSDI population which aged through '65 years old' that there are still more people living with HIV over 65 known to Medicare than to CDC in aggregates and overages impacted 70% of said volumes (Table 3). Race, Place_Race, and Age make up the remainder of errors, but their potential contains around 4% of volumes. Most likely, these errors are due to race attribution differences within series as well as Ryan White reporting children who are HIV- dependents of HIV+ beneficiaries (Table 3).

For Age aggregate, surveillance volumes being eclipsed services ends in 2014 for all services (Figure 1). Race and Race_Age have similar decreases, though the increase of Place_Race should be understood as the index gaining Ryan White Part A records in 2014, and not necessarily a decrease in surveillance completeness; the same is observed for Race_Age records (Figure 1). Death services data overtaking surveillance shows no sign of slowing and may be getting worse within domains over time (Figure 1). Improvements could be due to improvements to surveillance (CDC achieved all states and territories reporting in 2012) or because so many Americans have HIV at surveillance year 2014 that a surveillance excess within (sparse) services age groups becomes difficult to observe explicitly.

Within distributions of member domains within aggregates (i.e. % of a given race within races for a given year and program) off linear patterns within the aggregate distribution plotted against the surveillance distribution may indicate incomplete services case capture or service utilization bias. Within Medicare, Age and Race_Age departures from linearity should be understood as an artifact of the program which includes people living with HIV who are 65+ which is not representative of the age distribution of HIV in CDC surveillance (Figure 2). All other assessed distributions indicate potential SUSK cases or service bias within the index regions where surveillance exceeds service cases (dots outside grey bands) (Figure 2).

Age presents as linear though the error is heavily driven by Medicare (Figure 2). This could be because of the number of years in 65+ relative to other five-year age groups; and disambiguation of 65 from other implicit age groups may improve results (i.e., 66-70, 71-75). Race presents with off-linear domains though they are not radically off-linear and plotting is fairly close to local area expectations (Figure 2). Age_Death volumes have wide distributions but are not markedly out of local area either (Figure 2). SSDI and Medicaid have more 'out of locality' distributions and notable domains approaching 1 on the x axis, where greater than 1 is a surveillance excess (Fig 2). Such programs likely have SUSK cases, as Medicaid and SSDI are not universal health programs but are niche providers of low-income and completely disabled, no-income individuals respectively. Race_Age shows locally consistent distributions, not unlike race while Place_Race details many aggregates off trend (Figure 2). These uncertainties may be due to differences in race attribution or SUSK cases in the model. The presence of all programs having aggregates out of local range in Place_Race may suggest SUSK uncertainty for specific states and races in the US. Several race aggregates are overrepresented in Race_Death relative to their share in surveillance (Figure 2). This difference could be due to CMS capturing different people living with HIV within Race_Death, or differences in how CDC and CMS attribute race to individuals (especially 'Hispanics').

The deep learning models revealed anomalies worthy of further consideration. The models themselves performed reasonably well despite their 'ad-hoc', or untuned nature with r2 ranges between .82 and .98 (Table 4). Towards anomalies in predicting services from surveillance, program type is high value in resolving uncertainty, as to be expected when predicting services (Table 5). Differentiating Hispanic and knowing the ages of people living with HIV over 65 was also high value in predicting surveillance (Table 5). Specific aggregates can be identified as having greater use in predicting services vs surveillance when the 'diff; measure is considered (Figure 3).

Towards predicting surveillance, nine out of the top fifteen most important index terms were racial. This means that to understand the HIV surveillance data the race of the service seeker was more important than other model features (geography, age). This could be due to distinct epidemics within race groups, as Alaskan Indian, Native American (AIAN), Hispanic, Asian and Black all appear in the top fifteen variable importance when predicting surveillance (Table 5). Their presence could also be due to race specific service seeking behaviors. Record origins for Ryan White part A is also ranked in the top fifteen unlike Medicare, Medicaid and SSDI. Towards candidate SUSK measures, the difficulty in reconciling (the importance in known) race and age groups, as well as Place_Race pairs like New York and California Hispanics and South Carolina Blacks potentially suggests dimensionalities in the services-surveillance relationship worthy of future considerations (Table 5). Further, the ability to apply a coherent measure to aggregates of HIV populations over time creates the possibility of intentional investigations for local areas and populations in future deep learning models.

Interestingly, understanding surveillance for the deep learner requires the consideration of specific, high ranking domain volumes (California, Hispanic, New York, Hispanic, South Carolina Black). These domain volumes may be opportunities to improve surveillance in the real world. Because the model can return geographic locations, geo-specificity improvements (home zip-code) may sharpen future evaluations. Ultimately, the models demonstrate that when services are compared to surveillance, disagreement can be ascertained even when surveillance exceeds services.

### 4.2    Ethics concerns with large scale HIV data

Reviewers raised welcome concerns with the close fit of this study to emerging ethical issues in HIV research(35–37). Concerns with HIV status disclosure and potential population level consequences of integrating HIV surveillance and services data have also been raised publicly by advocates(38). The inclusion of a brief discussion about the ethics of HIV data reuse seemed appropriate and timely given the above study results as well as the continued emergence of similar data integration efforts (39–43).

Ethics concerns are traditionally resolved in bedside research using 'ethics methods' like informed consent or supporting patient participation in research design. These bedside ethics methods, not unlike biostatistics or chart reviews are not appropriate in large scale data work for several reasons. Given the scale of surveillance and services data used in this study re-identifying and consenting (millions of) people is impractical and not allowed under the data use agreement for this study. The labor requirements and third-party disclosure risks of such an effort would be extreme. The utility of a 'representative committee' is also superfluous in this case. No selection of people living with HIV could be meaningfully representative given the volume of persons observed in this national, retrospective study(44). As the big data turn has necessitated a re-writing of statistical methods to support data work at scale, it should also spur a re-writing of ethical methods with large scale data in mind. Any meaningfully successful big data turn should include an explicit bright line between acceptable and unacceptable reuse of data. Regrettably, no bright line currently exists except one borrowed from bedside medicine. Optimistically there is active debate in choice circles and a new standard may be forthcoming (45).

The deeper concern at the crux of ethics and large-scale HIV data work is the potential for population level consequences. If we can end the epidemic without centering the infected as (only) sources of infection or centering susceptible (high risk) individuals as sources of seroconversion in favor of some third, less stigmatizing approach remains to be seen. The first two approaches have been tried and to be fair, have failed to produce a knowledge that ends the epidemic. Studies should consider if kinds of stigmatization in knowledge production begets policies and practices which further accelerates HIV transmission(46).

As a non-bedside ethics question, the inherent utility and outcomes of HIV surveillance and HIV services remains under examined. They are assumed to be inherently good or at least more important than personal rights discourse or privacy. The dose-to-response relationship between surveillance and services to end the pandemic has been inferred by many, especially the 90-90-90 goals(4,8,47,48). Assumptions about surveillance and services mediating HIV transmission are not without their learned detractors(8,47–49). What we want surveillance and services to do may reflect the lack of investigation into the natural history of disease in an administrative context.

The data in our study follows the industry standard of only publishing non-identifiable information. We further only draw conclusions from non-identifiable information. In our current moment, bedside rules are being used to evaluate HIV data research harms. Most likely future ethicists will consider the 'bedside standard' insufficient but it is the current standard of practice (50). By practicing with transparency in an open journal there is at least hope that a bright line could emerge and form from a diversity of concerns with HIV data research practices.

## 5      Limitations

Opportunities to assess increases and decreases in case level uncertainty would be best learned from a 'complete' ground truth data set with consistent aggregation, which does not exist. CMS cases were considered retrospectively HIV+ following billing HIV on an inpatient or two outpatient claims over two years. CMS does not know when patients contracted HIV and the consistency of the CMS case distributions over time suggests that seroconversion after enrolling in CMS is perhaps rare.

The breadth of the 65+ age group may be concealing surveillance-services disagreement. The disambiguation of age group within 65+, especially for the Medicare population most likely reveals individuals who contract HIV very late (70's-90's) in life being conflated with people living with HIV who contract HIV in their 40s and 50s who then survive to enroll into Medicare as a prior-positive at age 65. Disambiguation of aggregates may add better model performance and improve identification of SUSK aggregates. Sharpening geo-specificity may also improve model utility.

## 6      Conclusion

Cases 'unknown' to surveillance can be discovered from existing services data where service cases exceed surveillance cases within aggregate and calendar year. When surveillance cases exceed services within aggregate and calendar year, modeling approaches may help in identifying suspect surveillance volumes. If eradicating HIV from the United States requires meaningful, qualified federal case surveillance, federal case surveillance efforts should consider using federal services data.

**Acknowledgments**

This research was carried out by staff of the National Library of Medicine (NLM), National Institutes of Health, with support from NLM. The authors have no conflicts of interests. The authors received no specific funding for this work.